\documentclass[5p]{elsarticle}
\usepackage{booktabs,shortvrb}
\usepackage{amsmath,amsfonts,amssymb}%
\usepackage{booktabs,url}
\usepackage{algpseudocode,algorithm}

\makeatletter
    \def\ps@pprintTitle{%
      \let\@oddhead\@empty
      \let\@evenhead\@empty
      \let\@oddfoot\@empty
      \let\@evenfoot\@oddfoot
    }
\makeatother
\begin{document}

\begin{frontmatter}
\title{A Unifying Survey of Reinforced, Sensitive and Stigmergic \\Agent-Based Approaches for E-GTSP }
\author[1]{Camelia-M. Pintea}\ead{cmpintea@yahoo.com}
\address[1]{Technical University Cluj-Napoca Romania}

\begin{abstract}
The {\em Generalized Traveling Salesman Problem (GTSP)} is one of the ${\cal NP}$-hard combinatorial optimization problems. A variant of {\em GTSP} is {\em E-GTSP} where $E$, meaning {\em equality}, has the constraint: exactly one node from a cluster of a  graph partition is visited. The main objective of the {\em E-GTSP}  is to find a minimum cost tour passing through exactly one node from each cluster of an undirected graph. Agent-based approaches involving are successfully used nowadays for solving real life complex problems. The aim of the current paper is to illustrate some variants of agent-based algorithms including ant-based models with specific properties for solving {\em E-GTSP}.
\end{abstract}

\end{frontmatter}

\section{Introduction}
A large number of combinatorial optimization problems are  ${\cal NP}$-hard. Nowadays, approximation and heuristic algorithms are used widely in order to find near optimal solutions of difficult problems, within reasonable running time. Heuristics are among the best strategies in terms of efficiency and solution quality for complex problems.

{\em The Generalized Traveling Salesman Problem (GTSP)} introduced in \cite{Laporte1983} and \cite{Noon1991} is also a complex and difficult problem. A variant of {\em GTSP}, {\em E-GTSP} where $E$ means "equality" is named generally just {\it GTSP} in the current paper. In {\em E-GTSP} exactly one node from a cluster is visited. 

Several approaches were considered for solving the {\em GTSP}. In \cite{Fischetti2002} a branch-and-cut algorithm for {\em Symmetric GTSP} is described and analysed. In \cite{Cachiani2010} is shown one the most recent paper in this area. The paper proposes a multistart heuristic ($MSA$) which  iteratively starts with a randomly chosen set of vertices and applies a decomposition approach combined with improvement procedures. 

A random-key genetic algorithm ({\em rkGA}) for the {\em GTSP} is  described in \cite{Snyder2006}. The {\em rkGA} combines a genetic algorithm with a local tour improvement heuristic with the solutions encoded using random keys \cite{Snyder2006}. Another genetic algorithm approach for solving {\it GTSP} is described in \cite{Silberholz2007}. The state-of-art {\em GTSP} memetic algorithm, proposed in \cite{Gutin2010}, exploited a strong local search procedure together with a well tuned genetic framework. In \cite{Renaud1998} it is proposed an efficient composite heuristic for the {\em Symmetric GTSP}. The $GI^3$ heuristic has three phases. First is constructing the initial partial solution. It follows the insertion of a node from each non-visited node-subset and in the third phase is a solution improvement phase \cite{Renaud1998}. 

There are significant achievements in the area of local search algorithms for the {\em GTSP}. In \cite{Karapetyan2012} is provided an exhaustive survey of {\em GTSP} local search neighbourhoods and proposed efficient exploration algorithms for each of them. Another effective {\em GTSP} local search procedure \cite{Karapetyan2011} is an adaptation of the well known Lin-Kernighan heuristic. A hybrid {\em ACS} approach using an effective combination of two local search heuristics of different classes is introduced in \cite{Reihaneh2012}.

{\em GTSP} has several applications to location and telecommunication problems. More information on these problems and their applications can be found in \cite{Fischetti1997,Fischetti2002,Laporte1983}. Other applications are in routing problems \cite{Pintea2011, Pop2009}. Hybrid heuristics are valuable instruments for solving large-sized problems. That is why several heuristics, including variants of ant-based algorithms are improved using different techniques. Some features of agents are involved as:  the level of sensibility, direct communications, the capability to learn and stigmergy.

Based on one of the best {\it Ant Colony Optimization} techniques, {\em Ant Colony System (ACS)} \cite{Dorigo1996}, in \cite{Pintea2006} was first introduced, {\em ACS} for solving {\em GTSP}. Using some {\em MAX-MIN} Ant System's \cite{Stutzle1997} features  and some new updating rules an reinforced {\it ACS} algorithm for {\em GTSP} was introduced in \cite{Pintea2006}. Computational results are reported for several test problems. The proposed  algorithm was competitive with already proposed heuristics for the {\em GTSP}. Several new heuristics involving agents properties were also introduced: {\it Sensitive Ant Colony System (SACS)}, {\it Sensitive Robot Metaheuristic (SRM)} and {\it Sensitive Stigmergic Agent System (SSAS)}. There are used two type of sensitive heuristics for solving {\em GTSP}. 

{\em Sensitive ACS (SACS)} \cite{Chira2007a} heuristic uses the sensitive reactions of ants to pheromone trails. Each agents is endowed with certain level of sensitivity allowing different types of responses to pheromone trails. The model involves search exploitation and search exploration in order to solve for complex problems. Numerical experiments illustrated in \cite{Chira2007a} shows the potential of the {\em SACS} model.

{\em Sensitive Robot Metaheuristic (SRM)} \cite{Pintea2008} uses virtual autonomous robots in order to obtain improved solutions of {\em SACS}. In {\it SSAS} \cite{Chira2007b} the agents adopt a stigmergic behaviour in order to identify problem solutions and have the possibility to share information about dynamic environments improving the quality of the search process. Using an {Agent Communication Language (ACL)} \cite{Wooldridge2005, Russell2003} the agents communicate by exchanging messages. This information obtained directly from other agents is important in the search process. 

The paper is organized as follows. Section~\ref{sect:def} provides a description and a mathematical model of the {\em  Generalized Traveling Salesman Problem}. In Section~\ref{sect:techniques} are illustrated the proposed agent-based models. Comparative numerical results and statistical analysis for the agent-based techniques involved for solving {\it GTSP} are illustrated in Section~\ref{sect:dis}. The paper concludes with further research directions.

\section{The GTSP description }
 \label{sect:def}
The current section includes a description of the {\it Generalized Traveling Salesman Problem} including a mathematical model and its complexity.

\subsection{A mathematical model of GTSP} The mathematical model of {\em GTSP} follows. There is considered the complete undirected graph $G=(V,E)$ with $n$ nodes. The graphs edges are associated with non-negative costs. The cost of an edge $e=\{i,j\}\in E$ is denoted by $c_{ij}$. 

The generalization of {\em TSP} implies an existing partition of set $V$. The subsets of $V$ are called {\em clusters}. Let $V_1,...,V_p$ be a partition of $V$ into $p$ {\em clusters} For example: $V=V_1 \cup V_2 \cup ... \cup V_p$ and $V_l \cap V_k = \emptyset$ for all $l,k \in \{1,...,p\}$. A {\em tour} is a subset  of nodes such that the subset contains exactly one node from each cluster of the graph partition.

\textit{Definition 1:} The objective of the Generalized Traveling Salesman Problem is to find a minimum-cost tour.

In other words, {\em GTSP} has to find a minimum-cost tour, a subset $H$, with exactly one node from each cluster  $V_i$, $i\in \{1,...,p\}$. {\em GTSP} involves the following decisions.

\begin{itemize}
\item Choose a node subset $S\subseteq V$, such that $|S \cap
V_k | = 1$, for all $k=1,...,p$ 
\item Find a minimum cost {\it Hamiltonian} cycle $H$ in the subgraph of $G$ induced by $S$.
\end{itemize}

\textit{Definition 2:} The GTSP is called {\it symmetric} if and only if the equality $c(i,j)=c(j,i)$ holds for every $i,j \in V$, where $c$ is the cost function associated to the edges of $G$. 

The time complexity for an exact algorithm is $|V_{k_1}|O(m+n\log n)$. In the worst case the complexity is $O(nm+nlogn)$ \cite{Pop2007}. An accurate discussion about time complexity for the {\em Generalized Traveling Salesman Problem} is given in \cite{Karapetyan2012}.
\section{Agent-based approaches for solving GTSP}
\label{sect:techniques}
The following subsections will describe in detail the reinforced, sensitive, multi-agent hybrid sensitive and stigmergic agent-based approaches for solving {\it GTSP}. 

\begin{figure}[htbp]
 \centering
		\includegraphics[scale=0.3]{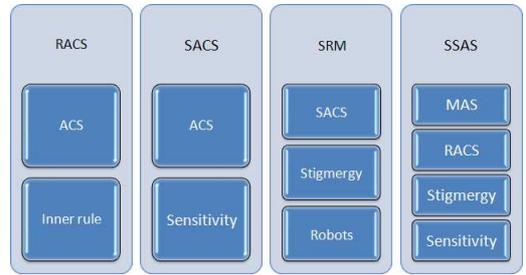}
	\caption{The successively development of the reinforced, sensitive and stigmergic agent-based models, starting with {\em Ant Colony System (ACS)}, using an reinforcement with inner-update rule in {\em Reinforcing Ant Colony System (RACS)}, involving sensitivity property for {\em Sensitive Ant Colony System (SACS)}, autonomous stigmergic robots for {\em Sensitive Robot Metaheuristic (SRM)}, {\em Multi-agent System (MAS)} and stigmergy in {\em Sensitive Stigmergic Agent System (SSAS)}}.
\label{fig:1}
\end{figure}

\begin{figure}[htbp]
 \centering
		\includegraphics[scale=0.3]{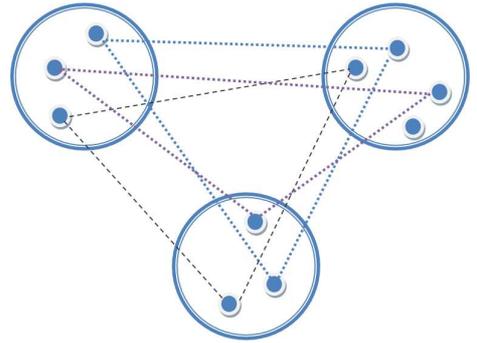}
\caption{A particular example of finding a minimum-cost tour spanning a subset of nodes such that the subset contains exactly one node from each cluster of the graph partition for the Equality Generalized Traveling Salesman Problem {\em E-GTSP}.}
\label{fig:2}
\end{figure}

In Figure \ref{fig:1} is an illustration of the successively development of the agent-based models and Figure \ref{fig:2} shows a particular example for the {\em E-GTSP}.

\subsection{Ant Colony System for GTSP}
\label{sect:acs}
The first  {\em Ant Colony Optimization} heuristic was {\em Ant System (AS)}. The algorithm was proposed in  \cite{Colorni1991,Dorigo1992}. It is a multi-agent approach used for various combinatorial optimization problems. 

The algorithm, as the entire {\it ACO} framework, was inspired by the observation of real ant colonies. 

In {\em AS} an artificial ant can find shortest paths between food sources and a nest. While walking from food sources to the nest and vice versa, the ants deposit on the ground a substance called pheromone. In this way a trail of pheromone is created. The real ants smells pheromone when choosing their paths.  The trails with the largest amount of pheromone is chooses. This feature employed by a colony of ants can lead to the emergence of shortest paths.  After a while the entire ant colony uses the shortest path. 

In {\em Ant System} are used artificial agents called artificial ants which iteratively construct candidate solution to an optimization problem. The solution construction is guided by pheromone trails and the specificity of each problem information.
\textit{Ant Colony System} ({\em ACS}) was developed to improve \textit{Ant System} making it more efficient and robust. \textit{Ant Colony System} for {\em GTSP} \cite{Pintea2006} works as follows.
\begin{itemize}
 \item  All \textit{m} ants are initially positioned on $n$  nodes chosen according to some initialization rule, for example randomly. Each ant builds an initial tour by applying a greedy rule. (see Algorithm 1.1.) 
\item The next node $j$, from an unvisited cluster is chosen, when the ant is in node $i$, depend on a variable $q$. The node $j$ is chosen with the maximal argument from equation Eq.~\ref{eqj} or using the probability from equation Eq.~\ref{eqprob}. While constructing its tour, the ant also modifies the amount of  pheromone  on the visited edges by applying the local
updating rule (Eq.~\ref{eqlocal}) (see Algorithm 1.2.).
\item After each step is computed the local best tour length (see Algorithm 1.3.)
\item Once all ants have finished their tour, the amount of pheromone on edges is modified again by applying the global updating rule. It is used the {\it Ant System} updating rule (Eq.\ref{eqglobal}.) knowing that an edge with a high amount of pheromone is a very desirable choice. The global updating rule follows in Algorithm 1.4.
\item The solution of the problem is the shortest tour found after a given number of iterations.
\end{itemize}
The already mentioned equations are detailed in Section \ref{sect:racs}. The sub--algorithms (Algorithm 1.1.--1.4.) and the {\it Ant Colony System} algorithm for {\it GTSP} follows.\\
\sffamily
\vspace*{0.15cm}

\begin{small}
\begin{tabular}{l}
\hline
{\bf Algorithm 1.1.} Initialization of GTSP \\
\hline
\vspace*{0.05cm}
1: {\bf forall } edges $(i,j)$ {\bf do}\\
2: \hspace*{0.25cm}$\tau_{ij}(0)=\tau_{0}$ \\
3: {\bf end for}\\
4: {\bf for }$k=1 $ to $m$  {\bf do}\\
5: \hspace*{0.25cm}place ant k on a randomly chosen node\\
6: \hspace*{0.25cm}from a randomly chosen cluster\\
7: {\bf end for}\\
8: build an initial tour $T$ using a Greedy algorithm\\
\hline
\end{tabular}
\end{small}

\newpage

\begin{small}
\begin{tabular}{ l }
\hline
{\bf Algorithm 1.2.} Construction of a tour for GTSP \\
\hline
\vspace*{0.05cm}
1: {\bf for} $k=1 $ to $m$  {\bf do}\\
2:  \hspace*{0.25cm}build tour $T^{k}(t)$ by applying \textbf{nc}-1 times\\
3:  \hspace*{0.5cm} {\bf if} ($q>q_0$) {\bf then}\\
4:  \hspace*{0.75cm}$j\in J^k_i$ is chosen with probability (Eq.~\ref{eqprob}) \\
5:  \hspace*{0.5cm}{\bf else} \\
6:  \hspace*{0.75cm}from an unvisited cluster choose node $j$ (Eq.~\ref{eqj})\\
7:  \hspace*{0.75cm}where $i$ is the current node\\
8:  \hspace*{0.5cm}{\bf end if}\\
9:  \hspace*{0.5cm}apply the local update rule (Eq.~\ref{eqlocal})\\
10: {\bf end for}\\
\hline
\end{tabular}
\end{small}

\vspace*{0.5cm}

\begin{small}
\begin{tabular}{ l }
\hline
{\bf Algorithm 1.3.} Compute a solution for GTSP \\
\hline

\vspace*{0.05cm}

1: {\bf for} $k=1 $ to $m$  {\bf do}\\
2: \hspace*{0.25cm}compute $L^{k}(t)$ of the tour $T^{k}(t)$\\
3: {\bf end for} \\
4: {\bf if}  an improved tour  {\bf then}\\
5: \hspace*{0.25cm} update $T^{k}(t)$ and $L^{k}(t)$\\
6: {\bf end if}\\
\hline
\end{tabular}
\end{small}

\vspace*{0.5cm}

\begin{small}
\begin{tabular}{ l }
\hline
{\bf Algorithm 1.4.} Global update rule for GTSP \\
\hline

\vspace*{0.05cm}

1: {\bf forall} edges $(i,j)\in T^{+}$  {\bf do}\\
2: \hspace*{0.25cm}update pheromone trails (Eq.~ \ref{eqglobal})\\
3: {\bf  end for}\\
\hline
\end{tabular}
\end{small}
\\
\vspace*{0.15cm}\\
 
\begin{small}
\begin{tabular}{ l }
\hline
{\bf Algorithm 1.} Ant Colony System for GTSP\\
\hline

\vspace*{0.05cm}

1: Initialization of GTSP\\
2: $T^{+}$ is the shortest tour and $L^{+}$ its length\\
3: {\bf repeat}\\
4: \hspace*{0.25cm}Construction of a tour for GTSP\\
5: \hspace*{0.25cm}Compute a solution for GTSP\\
6: \hspace*{0.25cm}Global update  rule for GTSP\\
7: {\bf until} end condition\\
8: {\bf return }$T^{+}$ and its length $L^{+}$\\ 
\hline
\end{tabular}
\end{small}\\
\rmfamily
\subsection{Reinforcing Ant Colony System for GTSP}
\label{sect:racs}

An {\em Ant Colony System} for the {\em GTSP} it is introduced and detailed in \cite{Pintea2006, PinteaPhDThesis}. In order to enforces the construction of a valid solution used in {\em ACS} a new algorithm called {\em Reinforcing Ant Colony System} {\em (RACS)} it is elaborated   with a new pheromone rule as in \cite{Pintea2005} and pheromone evaporation technique as in \cite{Stutzle1997}. 

Based on the mathematical model of {\em GTSP} from Section \ref{sect:def}, let  $V_k(y)$ be the node $y$ from the cluster $V_k$. The {\em RACS}
algorithm for the {\em GTSP} works as follows:

\begin{itemize}
\item Initially the ants are placed in the  nodes of the graph, choosing randomly the \textit{clusters} and also a random node from a  chosen cluster.
\item At iteration $t+1$ every ant moves  to a new node from an unvisited \textit{cluster} and the parameters controlling the algorithm are updated.
\item  Each edge is labelled by a trail intensity. $\tau_{ij}(t)$ is the trail intensity of the edge $(i,j)$ at time $t$. 

\bigskip

An ant decides which node is the next move with a probability that is based on the distance to that node, or the cost of the edge, and the amount of trail intensity on the connecting edge. The inverse of distance from a node to the next node is known as the \textit{visibility}, $\eta_{ij}$. 

\item At each time unit evaporation takes place in order to stop the intensity of pheromone on the trails. The rate evaporation is $\rho \in (0,1)$. 

A tabu list is maintained with the purpose to forbid ants visiting the same \textit{cluster} in the same tour. The ant tabu list is cleared after each completed tour.

\item In order to favour the selection of an edge that has a high pheromone value, $\tau$, and high visibility value, $\eta$ a probability function ${p^{k}}_{iu}$ is considered. ${J^{k}}_{i}$ are the unvisited neighbours of node $i$ by ant $k$ and $u\in {J^{k}}_{i},
u=V_k(y)$, being the node $y$ from the unvisited cluster $V_k$. 

The probability function is defined as follows:

\begin{equation}
\label{eqprob}
{p^{k}}_{iu}(t)= \frac{[\tau_{iu}(t)] [\eta_{iu}(t)]^{\beta}}
{\Sigma_{o\in {J^{k}}_{i}}[\tau_{io}(t)]\cdot [\eta_{io}(t)]^{\beta}},
\end{equation}

\indent where $\beta$ is a parameter used for tuning the relative
importance of edge cost in selecting the next node. 

${p^{k}}_{iu}$ is the probability of choosing $j=u$, where  $u=V_k(y) $ is the next node, if $q>q_{0}$, when the current node is $i$. 

If $q\leq q_{0}$ the next node $j$ is chosen as follows:
\begin{equation}
\label{eqj}
j=argmax_{u\in J^{k}_{i}} \{\tau_{iu}(t)
{[\eta_{iu}(t)]}^{\beta}\} ,
\end{equation}
\noindent where $q$ is a random variable uniformly distributed over $[0,1]$ and $q_{0}$ is a parameter similar to the temperature in simulated annealing, $0\leq q_{0}\leq 1$.

\item  The ants guides the local search by constructing promising solutions based on good locally optimal solutions. After each transition the trail intensity is updated using the inner correction rule from \cite{Pintea2005}.(see Algorithm 2.1.)

\begin{equation}
\label{eqinner}
\tau_{ij}(t+1)=(1-\rho)\cdot \tau_{ij}(t)+\rho \cdot \frac{1}{n \cdot L^{+}} .
\end{equation}
where $L^{+}$ is the cost of the current known best tour. In {\it ACS} \cite{Dorigo1996} for {\em GTSP} the local rule is :
\begin{equation}
\label{eqlocal}
\tau_{ij}(t+1)=(1-\rho)\cdot \tau_{ij}(t)+\rho \cdot \tau_0 .
\end{equation}

\item As in {\em Ant Colony System} only the ant that generate the best tour is allowed to \textit{globally} update the pheromone. The global update rule is applied to the edges belonging to the {\it best tour}. The correction rule follows.
\begin{equation}
\label{eqglobal}
\tau_{ij}(t+1)=(1-\rho) \tau_{ij}(t)+\rho  \Delta \tau_{ij}(t) ,
\end{equation}
\noindent where $\Delta\tau_{ij}(t)$ is the inverse cost of the best tour.
\item In order to avoid stagnation it is used the pheromone evaporation technique introduced in {\it $MAX-MIN$ Ant System }\cite{Stutzle1997}, if $\tau_{ij}(t)$ is over the $\tau_{max}$ value, as in equation \ref{maxmin}.

\begin{equation}
\label{maxmin}
 \mbox{if  }  (\tau_{ij}(t)>\tau_{ij}(t))  \mbox{ then }  \tau_{ij}(t)=\tau_{0} . 
\end{equation}

When the pheromone trail is over an upper bound $\tau_{max}$, the pheromone trail is re-initialized. 

The pheromone evaporation is used after the global pheromone update rule. (see Algorithm 2.2.)
\end{itemize}
The {\em RACS} algorithm (see Algorithm 2) computes for a given time a sub-optimal solution, the optimal solution if it is possible and can be stated as follows. Algorithm 1.2 and Algorithm 1.4 from Section~\ref{sect:acs} are modified and described further in Algorithm 2.1 and Algorithm 2.2.\\

\sffamily
\begin{small}
\begin{tabular}{ l }
\hline
{\bf Algorithm 2.1.} Reinforced construction of tours for GTSP\\
\hline
\vspace*{0.05cm}
1: {\bf for} $k=1 $ to $m$  {\bf do}\\
2: \hspace*{0.25cm}build tour $T^{k}(t)$ by applying \textbf{nc}-1 times\\
3: \hspace*{0.25cm}{\bf if} ($q>q_0$) {\bf then}\\
4: \hspace*{0.5cm}$j\in J^k_i$ is chosen with probability (Eq.~\ref{eqprob}) \\
5: \hspace*{0.25cm}{\bf else} \\
6: \hspace*{0.5cm}from an unvisited cluster choose node $j$ (Eq.~\ref{eqj})\\
7: \hspace*{0.5cm}where $i$ is the current node\\
8: \hspace*{0.25cm}{\bf end if}\\
9: \hspace*{0.25cm}apply the new local update rule (Eq.\ref{eqinner}) \\
10: {\bf end for}\\
\hline
\end{tabular}
\end{small}

\vspace*{0.5cm}

\begin{small}
\begin{tabular}{ l }
\hline
{\bf Algorithm 2.2.} Reinforced global update  rule for GTSP\\
\hline

\vspace*{0.05cm}

1:{\bf forall} edges $(i,j)\in T^{+}$  {\bf do}\\
2:\hspace*{0.25cm}update pheromone trails (Eq.~\ref{eqglobal}, Eq.~\ref{maxmin})\\
3:{\bf  end for}\\
\hline
\end{tabular}
\end{small}

\vspace*{0.5cm}

\begin{small}
\begin{tabular}{ l }
\hline
{\bf Algorithm 2.} Reinforcing Ant Colony System for GTSP\\

\hline
\vspace*{0.05cm}

1:  Initialization of GTSP\\
2:  $T^{+}$ is the shortest tour and $L^{+}$ its length\\
3:  {\bf repeat}\\
4:  \hspace*{0.25cm}Reinforced construction of a tour for GTSP\\
5:  \hspace*{0.25cm}Compute a solution for GTSP\\
6:  \hspace*{0.25cm}Reinforced global update  rule for GTSP\\
7:  {\bf until} end condition\\
8:  {\bf return }$T^{+}$ and its length $L^{+}$\\
\hline
\end{tabular}
\end{small}
 \rmfamily

\subsection{Sensitive Ant Colony System for GTSP}
\label{sect:kept}
The {\it Sensitive Ant Colony System (SACS)} for {\it GTSP} is based on the {\em Heterogeneous Sensitive Ant Model for Combinatorial Optimization} introduced in \cite{Chira2008}. {\it SACS} was introduced in \cite{Chira2007a}. 

In sensitive ant-based models there are used a set of heterogeneous agents (sensitive ants) able to communicate in a stigmergic manner and take individual decisions based on changes of the environment and on pheromone sensitivity levels specific to each agent. The sensitivity variable induce various types of reactions to a changing environment.

A good balance between search diversification and exploitation can be achieved by combining stigmergic communication with heterogeneous agent behaviour. 

Each agent is characterized by a pheromone sensitivity level, $PSL$ expressed by a real number from $[0, 1]$.  The transition probabilities from {\it ACS} model \cite{Dorigo1996} are changed using the $PSL$ values in a re-normalization process. The {\it ACS} transition probability is reduced proportionally with the {\it PSL} value of each agent in the sensitive ant-based approach \cite{Chira2008}.

Extreme situations of $PSL$ values are:
\begin{itemize}
\item When an ant is 'pheromone blind', meaning $PSL = 0$, therefore the ant  ignore  completely the stigmergic information 
\item When an ant has maximum pheromone sensitivity, meaning $PSL = 1$.
\end{itemize}
Low $PSL$ values indicate that a sensitive ant will choose very high pheromone levels moves. These ants are more independent and can be considered environment explorers and have the potential to discover in an autonomous way new promising regions. The ants with high $PSL$ values are able to intensively exploit the promising search regions already identified. The $PSL$ value can increase or decrease according to the search space encoded in the ant's experience.

In the {\it SACS} model for solving {\it GTSP} two ant colonies are involved. Each ant is endowed with a pheromone sensitivity level. In the first colony the ants have {\it small} $PSL$ values ($sPSL$) and the second colony with {\it high} $PSL$ values ($hPSL$). 

The $sPSL$ ants autonomously discover new promising regions of the solution space to sustain search diversification.  The sensitive-exploiter $hPSL$ ants normally choose any pheromone marked move. {\it SACS} for solving {\it GTSP} works as follows.

\begin{itemize}
\item As in {\em ACS} and {\em RACS}, initially the ants are placed randomly in the nodes of the graph.
\item At iteration $t + 1$ every $sPSL$-ant moves to a new node and the parameters controlling the algorithm are updated. 

When an ant decides which node from a cluster is the next move it does so with a probability that is based on the distance to that node and the amount of trail intensity on the connecting edge. At each time unit evaporation takes place. In order to stop ants visiting the same cluster in the same tour a tabu list is maintained. 

What differs from {\em ACS} and {\em RACS} models is the sensitivity feature.  The sensitivity level is denoted by $s$ and its value is randomly generated in $(0; 1)$. 

For $sPSL$ ants the sensitivity parameter $s$ is in $(0; s_0)$, where $s_0\in[0,1]$.

\item The trail intensity is updated \cite{Chira2007a}, using the local rule as following.
\begin{equation}
\label{eqlocalsacs}
 \tau_{ij}(t + 1) = s^2 \cdot  \tau_{ij}(t) + (1 - s)^2\Delta \tau (t) \cdot \frac{1}{n},
\end{equation}
where $n$ is the total number of the nodes.
\item The already mentioned steps are reconsidered by the $hPSL$-ant using the information of the $sPSL$ ants. For $hPSL$ ants $s$ values are randomly chosen in $(s_0; 1)$.
\item Only the ant generating the best tour is allowed to globally update the pheromone. The global update rule is applied to the edges belonging to the best tour. The correction rule is Eq.\ref{eqglobal}.
\end{itemize}
A run of the algorithm returns the shortest tour found. The description of the {\it SACS} algorithm for {\it GTSP} is shown in Algorithm 3.
\sffamily
\vspace*{0.25cm}

\begin{small}
\begin{tabular}{ l }
\hline
{\bf Algorithm 3.} Sensitive Ant Colony System for GTSP\\
\hline

\vspace*{0.05cm}

1: Set parameters, initialize pheromone trails\\
2: {\bf repeat}\\
3: \hspace{0.25 cm}Place ant k on a randomly chosen node\\
4: \hspace{0.25 cm}from a randomly chosen cluster\\
5: \hspace{0.25 cm}{\bf repeat }\\
6: \hspace{0.50 cm}Each sPSL-ant  build a solution (Eq.~\ref{eqprob},Eq.~\ref{eqj})\\
7: \hspace{0.50 cm}Local  updating rule (Eq.~\ref{eqlocalsacs})\\
8: \hspace{0.50 cm}Each hPSL-ant build a solution (Eq.~\ref{eqprob},Eq.~\ref{eqj})\\
9: \hspace{0.50 cm}Local updating rule (Eq.~\ref{eqlocalsacs})\\
10: \hspace{0.25 cm}{\bf until} all ants have built a complete solution\\
11: Global  updating rule (Eq.~\ref{eqglobal})\\
12: {\bf until} end condition\\
\hline
\end{tabular}
\end{small}
 \normalfont

\subsection{SRM for solving GTSP}

A particular technique, inspired from both {\em SACS} and involving autonomous robots is {\em Sensitive Robot Metaheuristic (SRM)}. {\em SRM} was introduced in \cite{Pintea2008}.

The model relies on the reaction of virtual sensitive autonomous robots to different stigmergic variables. Each robot is endowed with a distinct stigmergic sensitivity level. {\em SRM} ensures a balance between search diversification and intensification.
 
As it is detailed in \cite{Pintea2008}, a stigmergic robot action is determined by "the environmental modifications caused by prior actions of other robots". {\it Sensitive robots} are artificial entities with a {\em Stigmergic Sensitivity Level (SSL)} which is expressed by a real number in the unit interval [0, 1].

As it is in general for agents, here, in particular, robots with small {\em SSL} values are considered explorers of the search space and are considered independent sustaining diversification. The robots with high {\em SSL} values are exploiting the promising search regions already identified by explorers. The {\em SSL} values in {\em SRM} model increase or decrease based on the search space topology encoded in the robot experience.

Now something about the stigmergic robots involved in the process of solving a combinatorial optimization problem, including {\em GTSP}. 

Qualitative stigmergy \cite{Bonabeau1999,Theraulaz1999} means a different action \cite{Bonabeau1999,Theraulaz1999} and quantitative stigmergy is interpreted as a continuous variable which change the intensity or probability of future actions.

Because the robots have not the capability of ants to deposit chemical substances on their trail, a qualitative stigmergic mechanism is involved in {\em SRM}. These robots communicate using the local environmental modifications that can trigger specific actions. There is a set of so called "micro-rules" defining the action-stimuli pairs for a homogeneous group of stigmergic robots. These rules define the robots particular behaviour and find the type of structure the robots will create \cite{Bonabeau1999,Theraulaz1999}.  

In \cite{Pintea2008} the algorithm is used to solve a large drilling problem, a particular {\em GTSP} problem. In the following is a detailed description of the {\em SRM} for {\em GTSP}.
\begin{itemize}
\item Initially the robots are placed randomly in the search space. 
A robot moves at each iteration to a new node. The parameters controlling the algorithm are updated. 
\item The next move of a robot is probabilistically based on the distance to the candidate node and the stigmergic intensity on the connecting edge. In order to stop increasing  stigmergic intensity, evaporation process is invoked.  Also, is maintained a tabu list preventing robots to visit a cluster twice in the same tour. The stigmergic value of an edge is $\tau$ and the visibility value is $\eta$.

As in previous sections, ${J^{k}}_{i}$ is the unvisited successors of node $i$ by robot $k$ and $u\in {J^{k}}_{i}$. The {\em sSSL} robots probabilistically choose the next node.  $i$ is the current robot position. As in previous presented ant-based techniques the probability of choosing $u$ as the next node is given by \ref{eqprob}.

An autonomous robot could be in the team with high or in the team with low stigmergic sensitivity on the basis of a random variable uniformly distributed over $[0,1]$. Let $q$ be a realization of this random variable and $q_{0}$ a constant, $0\leq q_{0}\leq 1$. The robots with small stigmergic sensitivity {\em sSSL} are characterized by the inequality $q>q_{0}$ while for the robots with high stigmergic sensitivity {\it hSSL} robots $q\leq q_{0}$ holds.

A {\it hSSL-robot} uses the information given by the {\it sSSL} robots. {\em hSSL} robots choose the new node $j$ in a deterministic manner according to \ref{eqj}. The trail stigmergic intensity is updated using the local stigmergic correction rule:

\begin{equation}
\tau_{ij}(t+1)=q_{0}^2\tau_{ij}(t)+(1-q_{0})^2\cdot \tau_{0}.
\label{tauij2}
\end{equation}
\item Global updating the stigmergic value is the role of the elitist robot that generates the best intermediate solution. 

\bigskip

These elitist robots are the only robots having the opportunity to know the best tour found and reinforce this tour in order to focus future searches more effectively. This global updating rule is:

\begin{equation}\label{global22}
\tau_{ij}(t+1)=q_{0}^2 \tau_{ij}(t)+(1-q_{0})^2 \cdot \Delta \tau_{ij}(t) ,
\end{equation}

\noindent where $\Delta\tau_{ij}(t)$ is the inverse value of the best tour length. Furthermore $q_{0}$ is used as the evaporation rate factor.

\item An execution of the algorithm returns the shortest tour found. The stopping  criterion is given by a the maximal number of iterations ($N_{iter}$). 
\end{itemize}

The description of the {\em Sensitive Robot Metaheuristic} for solving the {\em GTSP} is illustrated further in Algorithm 4.

\sffamily
\vspace*{0.25cm}

\begin{small}
\begin{tabular}{ l }
\hline
{\bf Algorithm 4.} Sensitive Robot Algorithm for GTSP\\
\hline
\vspace*{0.05cm}
1: Set parameters, initialize stigmergic values of the trails;\\
2: {\bf repeat}\\
3: \hspace{0.15cm}Place robot k on a randomly chosen node\\
4:  \hspace{0.15cm}from a randomly chosen cluster\\
5:  \hspace{0.15cm}{\bf repeat }\\
6:  \hspace{0.30cm}Each robot incrementally builds a solution\\
    \hspace{0.7cm}based on the autonomous search sensitivity;\\
7:  \hspace{0.30cm}The sSSL robots probabilistically choose \\
\hspace{0.7cm}the next node (Eq.\ref{eqprob})\\
8:  \hspace{0.30cm}A hSSL-robot uses the information supplied by \\
    \hspace{0.7cm}the sSSL robots to find the new node j (Eq.\ref{eqj})\\
9:  \hspace{0.30cm}A local stigmergic updating rule (Eq.\ref{tauij2});\\
10: \hspace{0.15cm}{\bf until} all robots have built a complete solution\\
11: \hspace{0.15cm}A global updating rule is applied \\
\hspace{0.7cm}by the elitist robot (Eq.\ref{global22});\\
12: {\bf until} end condition\\
\hline
\end{tabular}
\end{small}
\normalfont
\subsection{Sensitive Stigmergic Agent System for GTSP}
\label{sect:ssas}
The {\it Sensitive Stigmergic Agent System for GTSP (SSAS)} introduced in \cite{Chira2007b} is based on the {\it Sensitive Ant Colony System (SACS)} \cite{Chira2007a} and {\it Stigmergic Agent System (SAS)} \cite{Chira2006}.

In \cite{Chira2006} was introduced the concept of stigmergic agents where agents communicate directly and also in a stigmergic manner using artificial pheromone trails produced by agents similar with some biological systems \cite{Camazine2001}. The novelty of {\it SSAS} is that the agents are endowed with sensitivity. Their advantage is that agents with sensitive stigmergy could be used for solving complex static and dynamic real life problems.

A multi-agent system ($MAS$) approach to developing complex systems involves the employment of several agents capable of interacting with each other to achieve objectives \cite{Jennings2001}. The benefits of $MAS$ include the ability to solve complex problems, interconnection and interoperation of multiple systems and the capability to handle distributed areas \cite{Wooldridge2005,Bradshow1997}. 

The {\it SSAS} model inherits also agent properties: autonomy, reactivity, learning, mobility and pro-activeness \cite{Wooldridge1999,Iantovics2009}. 

\bigskip

The agents are able to cooperate, to exchange information and can learn while acting and reacting in their environment. Agents also are capable to communicate through an agent communication language ({\it ACL}). 

If an agent has also sensitivity, stronger artificial \\pheromone trails are preferred and the most promising paths receive a greater pheromone trail after some time. Within the {\it SSAS} model each agent is characterized by a pheromone sensitivity level $PSL$ as in Section~\ref{sect:kept}.  The {\it SSAS} is using as in {\it SACS} two sets of sensitive stigmergic agents: with small and high sensitivity $PSL$ values. The role of sensitive ants from {\it SACS} is taken now, more generally, by sensitive-explorer agents, with small $PSL$ ($sPSL$ agents) and sensitive exploiter agents with high $PSL$ ($hPSL$ agents).

The $sPSL$ agents discover new promising regions of the solution space in an autonomous way,  sustaining search diversification. The $hPSL$ agents exploit the promising search regions already identified by the $sPSL$ agents. Each $PSL$ agent deposit pheromone on the followed path. Evaporation takes place each cycle preventing unbounded intensity trail increasing. The {\it SSAS} model for solving {\it GTSP} is described in the following. A run of the algorithm returns the shortest tour found.\\

\sffamily
\begin{small}
\begin{tabular}{ l }
\hline
{\bf Algorithm 5.} Sensitive Stigmergic Agent System for GTSP\\
\hline
\vspace*{0.05cm}
1:  {\bf Initialize} pheromone trails and knowledge base\\
2:  {\bf repeat}\\
3:  \hspace*{0.15cm}Activate a set of agents with various PSL\\
4:  \hspace*{0.15cm}Place each agent in search space\\
5:  \hspace*{0.15cm}{\bf repeat}\\
6:  \hspace*{0.3cm}Move to a new node each hPSL-agent (Eq.~\ref{eqprob}, Eq.~\ref{eqj})\\
7:  \hspace*{0.3cm}An agent send an ACL message\\
\hspace{0.7cm}about the latter edge formed\\
8:  \hspace*{0.15cm}{\bf until} all hPSL-agents have built a complete solution\\
9:  \hspace*{0.15cm}{\bf repeat} \\
10: \hspace*{0.3cm}Each sPSL-agent receive and use\\
\hspace{0.9cm}the information send by hPSL-agents\\
    \hspace*{0.9cm}or the information available in the knowledge base\\
11: \hspace*{0.3cm}Apply a local pheromone update rule (Eq.~\ref{eqinner})\\
12: \hspace*{0.15cm}{\bf until} all sPSL-agents have built a complete solution\\
13: \hspace*{0.15cm}Global pheromone update rule (Eq.~\ref{eqglobal})\\
14: {\bf until} end condition\\
\hline
\end{tabular}
\end{small}
\normalfont
\section{Evaluations of Agent-Based Algorithms for {\it E-GTSP}}
\label{sect:dis}
First some numerical experiments are illustrated in order to compare the already described algorithms. Based on these results and on the results from related papers are explained the advantages and disadvantages of the reinforced, sensitive and stigmergic  agent-based algorithms for {\it E-GTSP}. 

\subsection{Computational Analysis} In order to evaluate the performance of the already mentioned algorithms are used euclidean problems converted from {\em TSP} library \cite{tsplib}. In order to divide the set of nodes into subsets was used the procedure proposed in \cite{Fischetti1997} as in \cite{gtsplib,Gutin2010} and \cite{KarapetyanGTSP2012}. For this survey paper are used Euclidean problems of the Padberg-Rinaldi data set of city problems that can be obtained from the {\em GTSP} Instances Library \cite{Gutin2010}. 

In the related papers \cite{Chira2007a, PinteaPhDThesis, Chira2007b, Pintea2006} are detailed other numeric results. The algorithms were implemented in Java and tested on a $AMD Athlon$ 2600+, 333Mhz with 2GB RAM.  

\noindent {\it Parameters.} The parameters used for the agent-based approaches  are set as follows. 
\begin{itemize}
\item The initial value of all pheromone trails, $\tau_{0}=\frac{1}{n \cdot L_{nn}}$, the upper bound for the pheromone evaporation phase is considered $\tau_{max}=\frac{1}{1-\rho}\cdot \frac{1}{L_{nn}}$, where $L_{nn}$ is the solution  of \textit{Nearest Neighbor} algorithm (see \cite{Reihaneh2012}). 
\item Other values of the parameters are $\beta=5$, $\rho=0.5$, $q_{0}=0.5$ and ten number of ants for all considered algorithms. 
\item Besides the settings inherited from {\it ACS}, the {\it SACS} algorithm for {\it GTSP} uses an sensitivity parameter $s_{0}=0.5$. The sensitivity level of {\it hPSL} ants is considered to be distributed in the interval $(s_{0}, 1)$ while {\it sPSL} ants have the sensitivity level in the interval $(0, s_{0})$. In {\em SRM} the $SSL$ parameter is considered a random value in $[0,1]$.
\item It has been tested and observed that the best results are obtained by {\it SSAS} strategies assigning low {\it PSL} values for the majority of agents. $PSL$ is considered $0.01$ for all agents.
\end{itemize}

All the solutions of agent-based approaches are the average of five successively run of the algorithm, for each problem. The maximal computational time is set by the user, in this case ten  minutes.

In the following tables are compared the computational results for solving the {\em GTSP} using the {\em ACS}, {\em Reinforced ACS (ACS)}, {\em Sensitive Ant Colony System (SACS)} and {\em Sensitive Robot Metaheuristic (SRM)} and {\em Sensitive Stigmergic Agent System (SSAS)}. The columns in tables are as follows:

{\it Problem}: The name of the test problem. The digits at the beginning of the name give the number of clusters ($nc$); those at the end give the number of nodes ($n$).

{\it ACS, RACS, SACS, SRM, SSAS}: The gap of mean values after five runs, returned by the already mentioned agent-based algorithms. The gap is a percentage value computed as the difference between optimal and an algorithm solution, divided with the optimal solution.

\begin{table}\small\setlength{\tabcolsep}{4pt}
\centering
\caption{Agent-based approaches {em ACS}, {\em RACS}, {\em SACS}, {\em SRM} and {\em SSAS} comparative mean results}
\vspace{0.5cm}
\label{tab:2}
\begin{tabular}{lllllllll}
\hline
\noalign{\smallskip}
Problem & ACS	& RACS &	SACS &	SRM	 & SSAS   \\
\hline\noalign{\smallskip}
\noalign{\smallskip}
16PR76  & 0     & 0	  &0	&0	&0 \\
22PR107 & 0.06	& 0	   &0.01	&0.13	&0 \\
22PR124 & 0.30	& 0	   &0.14	&0.01	&0\\
28PR136 & 0.23	& 0	   &0.12	&0.05	&0\\
29PR144 & 1.47	& 0	   &0.04	&0.14	&0\\
31PR152 & 1.07	& 0.01	&0.53	&0.32 & 0\\
46PR226 & 2.82	& 0.21	&1.24	&1.44 & 0\\
53PR264 &2.76	  & 0.01	&0.54	&0.62	&  0\\
60PR299 & 4.25	& 0.53	&0.26	&0.24	&0.13\\
88PR439 & 39.19	& 4.72	&4.73	&5.86	&0.89\\
201PR1002&44.15 & 21.24 &22.20&18.15&16.30\\
 \hline
\end{tabular}
\end{table}
 \normalsize

\subsection{Statistical analysis: advantages and disadvantages} In Table~\ref{tab:2} are the mean values of five successively runs for each instance. For the smallest instances, with the number of clusters less than 40, each proposed algorithm have at least one optimal result. Between 40 and 60 all except {\em ACS} found at least once the optimal value. For a large number of clusters, over 60, the optimal value was never found, but the smallest value was found for {\em SSAS}. For the other algorithms the mean values is better than for {\em SSAS}.

The Expected Utility Approach \cite{Golden1984} technique is employed for statistical analysis purposes. The results of the test are shown in Table~\ref{table:22}.

Let $x$ be the percentage deviation of the heuristic solution and the best known solution of a particular heuristic on a given problem:

$$x=\frac{heuristic solution-best known solution}{best known solution}\times 100.$$

The expected utility function $euf$ can be expressed as:

$$euf = \gamma -\beta(1-\overline{b}t)^{-\overline{c}},$$

where $\gamma=500$, $\beta=100$ and $t=0.05$. 
$\overline{b}$ and $\overline{c}$ are the estimated parameters of the Gamma function. All values are translated with five units in order to use the current statistical analysis technique. There are considered $np=11$ problems for testing, the following notations are used in Table~\ref{table:22}:

$$\overline{x}=\frac{1}{np}\sum_{j=1}^{np}x_{j},
s^2=\frac{1}{np}\sum_{j=1}^{np}(x_{j}-\overline{x})^2, 
\overline{b}=\frac{s^2}{\overline{x}}, 
\overline{c}=(\frac{\overline{x}}{s})^2.$$
             
As indicated in Table~\ref{table:22}, {\em SSAS} model has Rank 1 (the last column in Table~\ref{table:22}) followed by {\em SRM} and {\em RACS} model. This result emphasizes that {\em SSAS} is more accurate compared to other techniques for the considered problem instances. {\em SSAS} is using the best features from each precedent algorithm.

\begin{table} \small
\centering
\caption{Statistical analysis results for compared agent-based models} \vspace{0.5cm}
\begin{tabular}{lllllll}
\hline    
&$\overline{x}$&$s^2$&$\overline{b}$& $\overline{c}$
&$euf$&Rk\\
\hline
\noindent$ACS$  & 1.0236  &8.7454	  & 8.5434  &0.1198	  & 393.0964& 5\\  
\noindent$SACS$ & 0.5420  & 1.4555 &2.6854   & 0.2018 &397.0472 & 4\\
\noindent$RACS$ & 0.4858  &1.3628  & 2.8051  & 0.1732 &397.3482	 &3 \\ 
\noindent$SRM$  & 0.4989  &  1.0521&	2.1088   &0.2366	  & 397.3288	& 2\\
\noindent$SSAS$ & 0.3149  &0.7974	  &2.5322   &0.1244  & 398.3022& 1\\ 
\hline
\end{tabular}
\label{table:22}
\end{table}

{\em Ant Colony System} shows once again the stability of the model introduced by \cite{Dorigo1992} and developed for {\em GTSP} in \cite{Pintea2006,PinteaPhDThesis}. As we can see from Table \ref{tab:2}, {\em RACS} performs on the small instances obtaining for many instances the optimal solutions for all execution of the algorithm.    

Sensitivity involved in {\em ACS} have the ability to identity good solutions for several instances and some optimal solutions too even for medium and large size instances.    

The autonomus stigmergic robots from {\em SRM} seems to have good results and have chances to be improved regarding the parameter values, execution time, may be using hybrid techniques or involving Lin-Kernighan algorithm and its variants \cite{Karapetyan2011, Reihaneh2012}. Another way to improve {\em SRM} could be making the robots working full parallel in inner loop of the algorithm.
 
$SSAS$ reports better running times for best values compared to the results others models suggesting the benefits of the model heterogeneity in the search process. The {\it SSAS} model can be improved in terms of execution time and using different values for parameters. Other improvements involves an efficient combination with other algorithms or the capability of agents working in parallel.

Each complex combinatorial optimization problem has his own particularities, therefore these biological inspired techniques should be tested and used further the agent-based metaheurisic with the best results. The introduced techniques could be also used for hybrid algorithms on improving classification techniques \cite{Parpinelli02datamining,Stoean2009}. Hybrid algorithms using these agent-based models have the chance to solve different real life {\em NP}-hard problems.

\section{Conclusion}
Several agent-based algorithms are involved for solving the equality {\em Generalized Traveling Salesman Problem}. Agents properties as autonomy, sensitivity, cooperation and $ACL$ language are strongly implied in the process of finding good solutions for the specified problem. The advantages of the reinforced, sensitive and stigmergic agent-based methods are the computational results, good and competitive with the existing heuristics from the literature. Some disadvantages are the multiple parameters used for the algorithms and the high hardware resources requirements.

\nocite{*}

\section{Acknowledgement}\noindent{\small I would like to express my appreciation to Daniel Karapetyan for his very helpful comments and valuable advices.}

\end{document}